# Towards Responsible AI for Financial Transactions

Charl Maree[1]
Center for AI Research
University of Agder
Grimstad, Norway
charl.maree@uia.no

Jan Erik Modal
SpareBank1 Development
SpareBank1 Alliance
Oslo, Norway
jan.erik.modal@sparebank1.no

Christian W. Omlin
Center for AI Research
University of Agder
Grimstad, Norway
christian.omlin@uia.no

*Abstract*—The application of AI in finance is increasingly dependent on the principles of responsible AI. These principles – explainability, fairness, privacy, accountability, transparency and soundness form the basis for trust in future AI systems. In this study, we address the first principle by providing an explanation for a deep neural network that is trained on a mixture of numerical, categorical and textual inputs for financial transaction classification. The explanation is achieved through (1) a feature importance analysis using Shapley additive explanations (SHAP) and (2) a hybrid approach of text clustering and decision tree classifiers. We then test the robustness of the model by exposing it to a targeted evasion attack, leveraging the knowledge we gained about the model through the extracted explanation.

## I. Introduction

AI is becoming increasingly more omnipresent in the financial industry, with applications in customer interaction, investor services, fraud detection, customer relationship management and anti money laundering [1]. There exists enormous business potential for advanced analytics and economic modelling. Customer relations can be improved through innovative services in the form of digital financial advisors or personal assistants. Personal assistants such as Google Alexa, Apple Siri and Google Assistant have been developed for many applications.

Currently, chatbots are the primary interface for digital assistants in finance. In the future, digital financial assistants will move beyond question answering and play a more active role in wealth management, smart payment solutions and credit and insurance management [1].

The need for the responsible application of AI in finance has been highlighted [1] [2] [3], and there is increased interest in responsible AI [4].

This study considers a financial transaction classification model. Electronic financial transactions are classified into categories, such as "groceries", "transportation", "savings", etc. The transactions are retrieved from the database of a major Norwegian bank and offer a good representation of the actual spending habits of customers. In Scandinavia, cash represents less than 5% of the total money supply [5] and is declining [6]. Electronic transactions therefore capture a significant portion of customer spending. The classifications made by the model of interest will, in the future, be used to develop a series of value adding products for customers, with the end goal of developing a digital financial advisor. As the basis for future work, it is important that the transaction classifier be implemented in accordance with the principles of responsible AI. In this study, we address one of the core principles of responsible AI, namely explainability.

The aim of this study is therefore to (1) identify the salient features of the transaction classification model and (2) extract an explanation for the function of the model, i.e. the rules that govern the model. We also illustrate vulnerability of the current financial transaction model to perturbations of the salient inputs. We achieve these goals in a hybrid approach where (1) we determine the feature importance using Shapley additive explanations (SHAP), (2) we generate explanations using a combination of clustering and decision trees and (3) we show model susceptibility to adversarial examples, leveraging knowledge from the explanation.

In Section II, we discuss the concept of responsible AI. We then briefly review related work in Section III. We describe our transaction classification model in Section IV and we present the theory behind the methods used for extracting an explanation. Section V describes the methodology for our experiments, and we discuss the results in Section VI. The paper closes with a summary, conclusions and directions for future research in Section VII.

## II. Responsible AI

Responsible AI provides a framework that focuses on ensuring the ethical, transparent and accountable use of AI technologies in a manner consistent with user expectations, societal laws and norms. It can guard against the use of biased data or algorithms, ensure that automated decisions are justified and explainable, and ensure user trust and individual privacy.

The principles of responsible AI can generally be summarized as fairness, privacy, accountability, transparency and soundness; however, no consensus exists on either a definition or measures for their quantification [4].

ML algorithms tend to adopt the bias present in the training data. This could translate into discrimination, e.g. credit rating according to postal codes [7], which violates the standards of fairness. AI systems can potentially use personal information in ways that intrude on individual privacy [8] by collecting and relating data that then becomes a commodity beyond the individual's knowledge or control. Accountability ensures that the system operator can be held liable for any adverse effects or consequences of the actions of AI systems; it does not necessarily remove bias. The imperative of AI transparency demands explainability and interpretability of AI systems, as well as data provenance. Explainability provides an accurate proxy or symbolic representation of the AI system whereas interpretability explains a model's predictions in human understandable terms, e.g. in relation to the input features. Explainability does not automatically imply interpretability [4]. Trustworthy AI systems must be reliable and accurate, behave predictably, and operate within in the boundaries of applicable rules and regulations. This also implies robustness and security against attacks such as

[1] Strategy Innovation and Development, SpareBank 1 SR-Bank ASA, Norway.

This research was partially funded by a grant from The Norwegian Research Council; project nr 311465

poisoning or evasion, as demonstrated in [9]. These core principles of responsible AI must be equally weighted in any responsible AI application.

III. RELATED WORK

Interpretable models require mitigation of their complexity; an explanation of an AI system may have high fidelity and accuracy, but it may be incomprehensible to humans. There is a common perception about the existence of a trade-off between model interpretability and performance [4]; the work reported in [10] addresses this issue. It unifies six existing methods; which lack certain desirable properties: (1) local accuracy, which requires the explanation model to at least match the output of the target model for some simplified input; (2) missingness, which requires features with zero values to have no attributed impact; (3) consistency, which states that if a model changes such that some simplified input's contribution does not decrease, then that input's attribution should increase or remain the same, irrespective of the other inputs.

The six unified methods are (1) local interpretable model explanations (LIME), which explains model predictions based on local approximations of the model around a given instance; (2) deep learning important features (DeepLIFT), which measures the change in model output resulting from changing a given input value to a reference value; (3) layer-wise relevance propagation, which estimates feature relevance from the changes prediction similar to DeepLIFT but uses a different underlying mechanism; (4) Shapley regression values, which calculate feature importance for linear models by retraining the model on different subsets of the features; (5) Shapley sampling values, which approximate the effect of removing a variable from the model by integrating over samples from the training set and (6) quantitative input influence, which addresses more than just feature importance, but that independently proposes sampling approximation which is nearly identical to Shapley values.

In general, calculating the exact SHAP values is a computationally impractical problem. SHAP unifies the insights from methods 1-6 to approximate them (see Section IV.B). In [11], the authors apply SHAP in order to explain the predictions of a non-linear model on a financial time-series. They reveal the salient features and show which features are responsible for predicting a given class of output. They show how SHAP values can be used to improve prediction accuracy by assessing the usefulness of adding additional data.

Once we have identified salient features, we intend to simplify the input space by means of clustering. In [12], the authors identify salient features, then use the most important feature to reduce model complexity through clustering of the input space; they then fit a unique decision tree on each cluster. The resulting small decisions trees are more compact and thus more interpretable than a single larger tree. In [13], a dataset is clustered in order to improve the performance of a decision tree classifier. The idea is that many smaller classifiers are more elastic in terms of underlying algorithms and parameters, compared to a single, larger classifier. The authors report a 40% improvement in classification performance using this method.

IV. CLASSIFICATION MODEL AND EXPLANATION

Our target system is a transaction classification system which is currently in production and receives between 10 and 1500 requests per second. A typical request has about 100 transactions and processing time for requests increases linearly with the number of transactions. Processing time is typically between 2ms and 50ms.

In this section, we discuss the features used in the target model and give a short overview of the target model. We then introduce the methods extracting an explanation.

A. Feature Encoding and Target Model

The features in the dataset include categorical, numerical and text attributes. The target model is a series of two opaque models: a word2vec encoder followed by a deep neural network (DNN). In the first model, the transaction text is encoded into a vector representation

$$X_t = \{X_t^i\}, i \in \{1, \ldots, n\} \subset N \quad (1)$$

where $N$ is the number of features in the feature space, $n$ is the dimensionality of the vector representation of the text, i.e. the product of the size of embedding vector $k$, and the number of words in the text $l$, i.e. $n = k \times l$. This vector is concatenated with one-hot encodings of the transaction code $X_c$ and day of week $X_d$, normalized transaction amount $X_a$ and customer age $X_g$ as well as binary series representing whether the transaction amount is negative or positive (payment vs deposit) $X_d$ and whether the amount includes cents $X_e$. This concatenated dataset $X$, which is sent into a DNN is formally represented by:

$$X = \{X_t, X_c, X_d, X_a, X_g, X_d, X_e\} \quad (2)$$

The model is a classification net, producing a probability distribution $Y_i \in Y$, $i \in \{1, \ldots, m\}$ where $m$ is the number of output classes.

The training set was labelled using a mixed technique of defined rules and manual labelling. The rules did not accurately classify all transaction; misclassified transactions had to be hand labelled.

B. Salient Feature Extraction using SHAP

Shapley additive explanations (SHAP) [10] is based on the collaborative game theory method, Shapley values [14]. It clarifies the prediction of an instance $x \in X$, where $X$ is the set of all instances, by computing the contribution of each input feature $x_i \in x$, $i \in \{1, \ldots, N\}$ where $N$ is the number of features in the dataset. SHAP values assign weights to each feature cluster, where a feature cluster can be either a single feature, e.g. in numeric data, or a group of features, e.g. several words in a sentence. SHAP uses these weights in an additive linear model to explain the overall contribution of all features, thus elegantly blending elements from Shapely values [14], LIME [15] and others.

In [10], the authors define a given explanation model $g$ as

$$g(z') = \phi_0 + \sum_{j=1}^{N} \phi_j z'_j \quad (3)$$

where $z' \epsilon \{0,1\}^N$ is the feature space vector indicating the presence of each feature, $N$ is the size of the feature space and SHAP values $\phi_j \epsilon \mathbb{R}$ is the individual feature contribution for a feature $j$. The feature space refers to a simplified feature space that maps to the original feature space through a mapping function $z = h_z(z')$. The individual feature contributions $\phi_j \in \mathbb{R}$ are estimated using the collaborative game theory approach Shapley [14].

Shapley explores a game where the prediction of a model $f(x)$ is seen as the result, or payout, of the game. The individual features $x_i \in x$ are the players. The goal is to determine the contribution that each player has to the payout. Shapley determines how to fairly distribute the payout among the players through comparison of the model outputs for different coalitions of feature values. Feature coalitions are made by randomly sampling values from the feature space, i.e. a coalition is a fictitious instance $x' \notin X$, where feature values of the instance $x_i' \in x$ are drawn randomly from the feature space. The Shapley value of a feature is defined as the average change in the prediction $\Delta \hat{f}(x) = \hat{f}(x') - \hat{f}(x'')$ that a coalition $x'$ receives when a new feature value $x_i'' \in x$ joins the coalition.

*C. Text Clustering using DBSCAN*

Text is typically clustered using a spatial clustering algorithm, such as the density based spatial clustering algorithm with noise (DBSCAN) [16], [17]. It starts with a random instance and identifies all its nearest neighbors. Proximity to other instances is determined through a given distance measure, e.g. Euclidean, Hamming, Cosine, etc. If a point has a minimum of $minPts$ neighbors within a distance of $\epsilon$, then a new cluster is defined. The algorithm will also identify outliers that do not fall in any cluster as noise.

When text is represented as word vectors, through e.g. a word2vec encoder, the similarity between two sentences corresponds to the distance between the vectors. This is generally quantified as the cosine of the angle between the vectors [18] i.e. the cosine similarity. Given two sentences, the cosine similarity is defined as:

$$sim_c(t_i, t_j) = \frac{t_i \cdot t_j}{|t_i| \times |t_j|} \quad (4)$$

Where $t_i, t_j \in T$, are $n$-dimensional vectors in the term set $T = \{t_1, ..., t_n\}$ and $sim_c \in [0,1]$. When two terms are identical, the cosine similarity is 1 i.e. $sim_c(t_k, t_l) = 1, \forall\ t_k = t_l$.

DBSCAN can therefore be used with cosine similarity as a clustering method for texts.

## V. EMPIRICAL METHODOLOGY

*A. Data*

Throughout this study, we used an initial dataset of roughly 10 million financial transactions. These transactions were labelled using the target model and resampled without replacement to provide a more uniform representation of the labelled classes. The final dataset, $X$, had a cardinality of roughly 5 million transactions, i.e. $|X| \cong 5\ 000\ 000$.

*B. Explanation by Decision Trees*

Global surrogate modelling is a well-documented approach to model explainability [4]. In this study, we trained both a single decision tree and a random forest as global surrogates to explain the model. We used a random sample of 10% of the total dataset (about half a million transactions) for training, $X_{train} \subset X\ \wedge\ |X_{train}| = 0.1 \times |X|$, while testing was done on a randomly sampled set of 100 000 transactions, $X_{test} \subset X\ \wedge\ X_{test} \notin X_{train}\ \wedge\ |X_{test}| = 100\ 000$.

We used these train and test sets to fit a decision tree classifier and a random forest classifier with 50 individual trees. The performance and human understandability of the tree and forest were used as a baseline to compare with a hybrid clustering / decision tree approach discussed below.

*C. Feature Importance through SHAP Analysis*

We estimate the feature importance using SHAP [10]. The feature importance, $\phi_i$ was estimated for each input feature $x_i\ i \in \{1, ..., N\}$ where $N$ is the total number of encoded features. Note that due to encoding, $N > 7$ where 7 is the number of original features.

Equations (1) and (2) illustrate how the features are prepared, with equation (1) referring to the word vectors for the transaction text. SHAP values provide an estimate of the importance of individual features, $\phi_i \rightarrow X_t^i$; however, this is not useful when the feature of interest is a superfeature: $X_t = \{X_t^i\}, i \in \{1, ..., n\}$. In order to derive the importance of the superfeature $X_t$, we aggregate the SHAP values through addition [10]:

$$\phi_t = \sum_1^n \phi_i \quad (5)$$

*D. Explanation through Clustering and Decision Trees*

Having identified the most important feature, we clustered the data according to this feature. In Section VI, we show that the most important feature in classification is the transaction text $X_t$; we therefore used the DBSCAN algorithm with cosine similarity as the distance measure. We trained a set of $m$ superclusters, $c_i \in C, i \in \{1, ..., m\}$, where $m$ is the number of classes in the output $y_i \in Y, i \in \{1, ..., m\}$.

From these superclusters, we considered the individual words from the texts contained in each cluster. We created a list of keywords $k_i \in K$ for each supercluster $i$ by extracting unique words from each cluster. Stop words such as place and street names were removed from the keyword lists. Formally,

$$k_i \in K\ \wedge\ k_i \cap k_j = \emptyset$$
$$i, j \in \{1, ..., m\} \wedge\ i \neq j \quad (6)$$

The keywords were used as rules that associate a given transaction text with a given supercluster. For any given transaction text $t$, each word in the text $w \in t$ was given the opportunity to vote for a supercluster $c$; we considered the keyword list for each supercluster; if a word $w$ appears in the keywords list $k_i$, the word voted for supercluster $c_i$. The votes for all words were tallied and the supercluster was selected through majority vote. If no supercluster was found, i.e. no words appear in the keyword list, a default supercluster representing the class "other" was selected. We used shallow decision trees to filter out those instances that did not belong to the homogeneous majority. This is similar to the approach in [12] and [13]; we intended to simplify the final explanation while simultaneously attaining improved accuracy compared to a single large classifier.

*E. Model Robustness against Evasion Attacks*

In order to test the robustness of the model, we subjected the model to a targeted evasion attack, leveraging the newfound knowledge about the model. A successful adversarial attack therefore suggests not only a vulnerability in the model, but also a working knowledge of the model by the attacker.

The adversarial examples were generated by slightly perturbing existing instances, along the feature of highest importance, i.e. where the impact would be greatest. The perturbations therefore targeted the transaction text, $X_t$. The perturbed set of adversarial examples $X_{pert} \in X$ is therefore defined by:

$$x' = \{x_t', x_c, x_d, x_a, x_g, x_d, x_e\} \tag{7}$$

$$x \in X \land x' \in X_{pert}$$

Words from the texts were selected by matching the words with the keyword dictionary, $K$. If a word appeared in one of the keyword lists $k_i \in K$, then that word was replaced by a word from another list $k_j \in K$, where $i \neq j \land i, j \in \{1, \dots, m\}$.

## VI. RESULTS

The labelled set of transactions was divided into training (80%), validation (10%) and test (10%) sets. The trained model achieved a mean accuracy of 98.2%, with a 95% confidence interval of 0.04% in 20 experiments.

As a baseline to an explanation, we trained a decision tree and a random forest as surrogate models on data labelled by the DNN. The decision tree achieved an accuracy of 95.9%, while the random forest (with 50 estimators) achieved an accuracy of 96.7%. Both the single decision tree and the random forest had in excess of 50 000 nodes. Even though decision trees inherently explained the rules they have derived, they clearly do not provide interpretability in this instance.

### A. Feature Importance and Model Explanation

The results from the SHAP feature importance evaluation are clear evidence of the model's bias towards the text features. As seen in Fig. 1, the transaction text is largely responsible for the predictions. This is consistent with the importance of transaction text for the partial labelling of the original dataset.

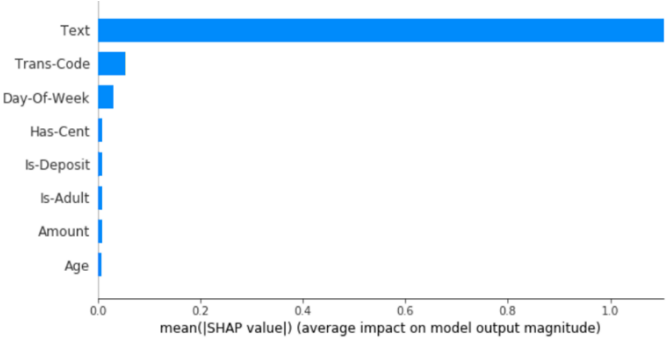

Fig. 1 The feature importance estimation by SHAP analysis shows that the transaction text is the most important feature for the transaction classification.

Knowing that the text is the most important feature for model classification is not an adequate explanation of the functioning of the model. To determine how the model uses the text, we used its vector representation in a clustering analysis; we used DBSCAN with the distance parameter $\epsilon = 0.07$. The intent was to train tight clusters. The result was a set of clusters with high homogeneity (95%) and a low percentage of noise (2%), with a total of 12 734 clusters. We then grouped the clusters using the labelled training data into $m$ superclusters. Fig. 2 shows a 2-dimensional representation of the supercluster for transactions relating to alcohol.

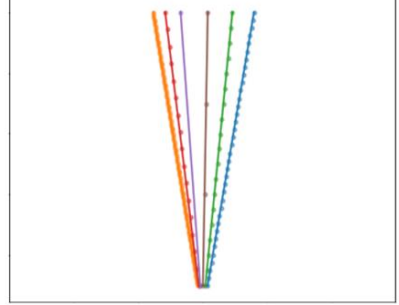

Fig. 2. Text vectors are clustered and grouped into superclusters. A 2-dimensional projection of the supercluster "Alcohol" is shown with each of the clusters containing several instances from the training set. The angles of the clusters shown in this plot are equal to the angles in the word2vec text embedding dimension.

Finally, we fit a small, interpretable decision tree to each supercluster with less than 100% homogeneity; the shallow decision tree provides the final separation and explanation. An example tree is shown in Fig. 3, for cluster number 10 relating to expenditure on kindergartens:

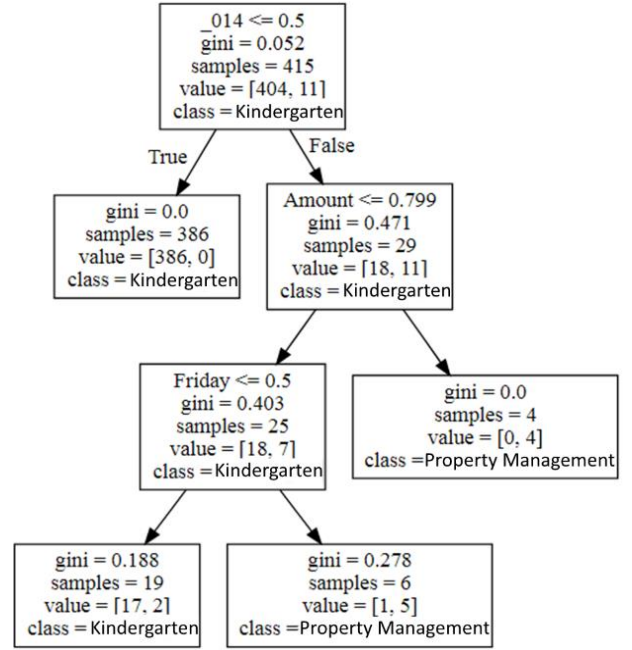

Fig. 3 The decision tree for supercluster 10 (kindergarten) makes the final distinction between transactions relating to kindergarten and those relating to property management. This is one of many trees, each relating to a single supercluster and separating instances observed in that supercluster during training.

In Fig. 3, the tree distinguishes between transactions relating to kindergarten and those relating to property management. The transaction code "014" is the most important feature in this classification, while the amount and day of week also play roles. In Fig. 4, we plot the feature importance for the decision tree shown in Fig. 3.

Fig. 4 shows that the transaction code is the most important feature. This correlates well to our previous estimated of the overall and average feature importance in the original model (Fig. 1). The SHAP feature importance coincides with the feature importance observed in Fig. 4.

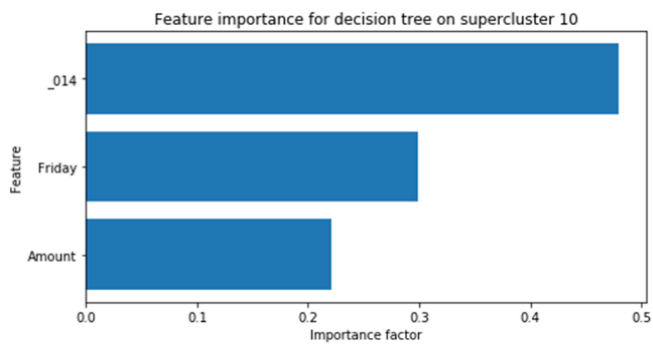

Fig. 4 Each decision tree supplies a feature importance estimate. The feature importance is shown for the decision tree of supercluster 10 (kindergarten).

The additive SHAP values allowed us to identify the transaction text as the dominant feature for transaction classification. Among the remaining features, the shallow decision trees identified the transaction code as the feature that filters transaction from heterogeneous clusters of the word2vec text embedding. In the large decision tree, the words from the transaction text were scattered throughout the nodes. It remains to be seen whether this is a general property.

We evaluated the fidelity of the explanations by comparing their prediction with those of the transaction model [19]. The explanation model typically made the same prediction as the transaction model for 98% of the labelled data.

To evaluate the transaction model's robustness to changes in the transaction text, we scored a perturbed dataset and found that the model prediction typically changed for 80% of transactions with a single word replaced. We then repeated the experiment with a new set of perturbed transactions, where we replaced more than one word; this typically resulted in 90% of the transactions being classified differently.

## VII. CONCLUSIONS AND DIRECTIONS FOR FUTURE WORK

In this paper, we introduced a transaction classification model which is the basis for future value adding products for banking customers, with the end goal of developing a digital financial advisor. It is thus imperative that the transaction classifier be implemented in accordance with two of the principles of responsible AI: explainability and robustness.

We found that decision trees and random forests derived from the transaction model may offer *explainability*, but their complexity (> 50 000 nodes) limits their *interpretability*.

We mitigated the complexity of the feature space by identifying the transaction text as salient. The text was then used to cluster the dataset, before fitting a small tree to each cluster where necessary. These decision trees offered improved *interpretability* as they were smaller and easier for a human to understand.

Finally, we briefly investigated the robustness of the model by subjecting it to an evasion attack. The large influence observed for text perturbations correlates well with our SHAP analysis which suggests a large model dependence on the text. We find that the model is vulnerable to changes in the transaction text. However, since vendors seldomly change their formulas for generating transaction texts and companies seldomly change their names, the text is mostly an immutable property of the transactions. This vulnerability is therefore deemed low risk for such transactions. In the case of bank transfer transactions where customers may enter free text, there could be risk of masking fraudulent or money laundering transactions. If the classifier was ever to be used to detect such transactions this would be a point to address.

ACKNOWLEDGMENT

We would like to thank the SpareBank 1 Alliance for useful discussions and SpareBank 1 SR-Bank for providing anonymized transaction data.